\documentclass{ieeeaccess}
\usepackage{cite}
\usepackage{amsmath,amssymb,amsfonts}
\usepackage{algorithmic}
\usepackage{graphicx}
\usepackage{textcomp}
\usepackage{cleveref}
\crefname{section}{Section}{Sections}
\Crefname{section}{Section}{Sections}
\crefname{figure}{Fig.}{Figs.}
\Crefname{figure}{Figure}{Figures}

\newcommand{\contextmatch}{\textbf{ConTeXT-match}}
\newcommand{\skillxl}{\textbf{Skill-XL}}
\newcommand{\newjobbert}{\textbf{JobBERT V2}}
\usepackage{pifont}
\usepackage{textcase}

\usepackage{multirow} 
\usepackage[table,xcdraw]{xcolor} 
\usepackage{booktabs} 
\usepackage{float}

\usepackage[utf8]{inputenc}

\usepackage{bm}
\makeatletter
\AtBeginDocument{\DeclareMathVersion{bold}
\SetSymbolFont{operators}{bold}{T1}{times}{b}{n}
\SetSymbolFont{NewLetters}{bold}{T1}{times}{b}{it}
\SetMathAlphabet{\mathrm}{bold}{T1}{times}{b}{n}
\SetMathAlphabet{\mathit}{bold}{T1}{times}{b}{it}
\SetMathAlphabet{\mathbf}{bold}{T1}{times}{b}{n}
\SetMathAlphabet{\mathtt}{bold}{OT1}{pcr}{b}{n}
\SetSymbolFont{symbols}{bold}{OMS}{cmsy}{b}{n}
\renewcommand\boldmath{\@nomath\boldmath\mathversion{bold}}}
\makeatother

\def\BibTeX{{\rm B\kern-.05em{\sc i\kern-.025em b}\kern-.08em
    T\kern-.1667em\lower.7ex\hbox{E}\kern-.125emX}}

\begin{document}
\history{Date of publication xxxx 00, 0000, date of current version xxxx 00, 0000.}
\doi{10.1109/ACCESS.2024.0429000}

\title{Efficient Text Encoders for Labor Market Analysis}
\author{\uppercase{Jens-Joris Decorte}\authorrefmark{1,2},
\uppercase{Jeroen Van~Hautte}\authorrefmark{1}, \\
\uppercase{Chris Develder}\authorrefmark{2}, \IEEEmembership{Senior Member, IEEE}, AND
\uppercase{Thomas Demeester}\authorrefmark{2}}

\address[1]{TechWolf, Ghent, Belgium}
\address[2]{Internet and Data Science Lab (IDLab), Ghent University – imec, Ghent, Belgium}
\tfootnote{This project was funded by the Flemish Government, through Flanders Innovation \& Entrepreneurship (VLAIO, project HBC.2020.2893) and under the ``Onderzoeksprogramma Artifici\"ele Intelligentie (AI) Vlaanderen'' program, and by TechWolf.}
\tfootnote{This work has been submitted to the IEEE for possible publication. Copyright may be transferred without notice, after which this version may no longer be accessible.}

\markboth
{J.J. Decorte \headeretal: Preparation of Papers for IEEE TRANSACTIONS and JOURNALS}
{J.J. Decorte \headeretal: Preparation of Papers for IEEE TRANSACTIONS and JOURNALS}

\corresp{Corresponding author: Jens-Joris Decorte (e-mail: jensjoris@techwolf.ai).}

\begin{abstract}
Labor market analysis relies on extracting insights from job advertisements, which provide valuable yet unstructured information on job titles and corresponding skill requirements.
While state-of-the-art methods for skill extraction achieve strong performance, they depend on large language models (LLMs), which are computationally expensive and slow.
In this paper, we propose \contextmatch, a novel contrastive learning approach with token-level attention that is well-suited for the extreme multi-label classification task of skill classification.  
\contextmatch\ significantly improves skill extraction efficiency and performance, achieving state-of-the-art results with a lightweight bi-encoder model.
To support robust evaluation, we introduce \skillxl\, a new benchmark with exhaustive, sentence-level skill annotations that explicitly address the redundancy in the large label space. 
Finally, we present \newjobbert, an improved job title normalization model that leverages extracted skills to produce high-quality job title representations.
Experiments demonstrate that our models are efficient, accurate, and scalable, making them ideal for large-scale, real-time labor market analysis.
\end{abstract}


\titlepgskip=-21pt

\maketitle

\section{Introduction}
\label{sec:introduction}

\PARstart{L}{abor} market analysis plays a central role in addressing global workforce challenges such as talent shortages, skill gaps, and fast-changing job requirements driven by technological advancement. 
Accurate insights into the skills demanded by employers inform a wide range of applications such as workforce planning and policymaking~\cite{dsjmaapplications,cunningham2016employer}.
In this context, job advertisements (job ads) have served as a valuable resource for understanding labor market trends~\cite{9517309}.
Job ads are published on a daily basis across industries and regions and contain fine-grained information about job titles and their respective skill requirements.
Although rich in information, job ads often use different terminology to refer to occupations and skills. 
Therefore, robust labor market analysis requires natural language processing (NLP) techniques to identify and normalize the information contained in job ads. 

To address this terminology challenge, the European Commission developed ESCO (European Skills, Competences, Qualifications and Occupations),\footnote{https://esco.ec.europa.eu/en} a multilingual classification system that serves as a common language for the labor market. 
ESCO provides a comprehensive taxonomy of close to 14,000 skills and over 3,000 occupations, enabling consistent analysis and comparison of job market data across time and different regions.
This standardized framework is particularly valuable for large-scale labor market analysis, as it allows for the systematic mapping of diverse job descriptions to a common reference point.

To leverage ESCO's standardized taxonomy effectively, two key tasks are essential: job title normalization and skill requirement extraction from job ads\,---\,skill extraction in short. 
The latter involves identifying the skills mentioned in a job advertisement and mapping them to their corresponding ESCO skill definitions, whereas job title normalization addresses the challenge of mapping diverse job titles to standardized ESCO occupations. 
Together, these tasks enable robust labor market analysis by transforming unstructured job descriptions into structured, comparable data points.

Research interest in skill extraction has grown steadily over the last decade~\cite{9517309}.
Recently, increased parsing capabilities and semantic understanding of large language models (LLMs) have been shown to achieve state-of-the-art results for skill extraction~\cite{clavieskill,d2024context}. 
Job title normalization has also been approached using LLMs, although with less convincing results~\cite{occcodingllm}.
However, insightful labor market analysis necessitates a large volume of job ads to be analyzed, making the need for efficient NLP models greater.
Lightweight models that achieve high performance without relying on vast computational resources can democratize access to labor market insights, enabling organizations and researchers to process large datasets cost-effectively and at scale. 

Benchmarks for skill extraction have traditionally been formalized as span labeling tasks, without linking the identified spans to respective skills in a taxonomy. 
This lack of normalization in skill extraction prevents robust analysis of aggregate job skill requirements because of e.g. synonyms. 
A few benchmarks do provide fine-grained skill labels, yet they are either aggregate labels on the full ad level (thus preventing sentence-level evaluation), or they are the result of post-annotation on sequence annotations, limiting the expressiveness of annotating implicit skills. 
Finally, while large fine-grained skill taxonomies are great for normalizing skill information and robust analysis, their large size often means that a degree of semantic overlap is present between some skill labels, making it more difficult to annotate the ground truth and evaluate the models in a robust manner.

In this work, we address these challenges through the following contributions:

\begin{enumerate}
    \item We introduce \contextmatch, a new contrastive learning approach with token-level attention, designed for extreme multi-label text classification. 
    We apply this to the skill extraction task by training a bi-encoder for sentence-level skill extraction. 
    The model outperforms all other skill extraction models with the same number of parameters, owing to a new token-level contrastive loss. 
    With just 109 million parameters, the model effectively closes the performance gap between encoder models and LLM-based skill extraction systems, achieving state-of-the-art results on most metrics. The model is made available online.\footnote{https://huggingface.co/TechWolf/ConTeXT-Skill-Extraction-base}
    \item We develop the skill extraction evaluation with our newly constructed e\textbf{X}haustive \textbf{L}abels (\skillxl) benchmark: a large manual annotation effort with a unique focus on annotating job ads with exhaustive labels, explicitly coding redundancy among skill labels in the annotations. 
    The dataset contains 111 job ads annotated on a sentence-by-sentence basis with a total of 8,471 skill labels, and is made available online.\footnote{https://huggingface.co/datasets/TechWolf/Skill-XL}
    \item We produce \newjobbert, a simplified and superior iteration of the earlier JobBERT model~\cite{jobbert} for job title normalization, which achieves results on par with those of complex state-of-the-art methods. \newjobbert\ is available online.\footnote{https://huggingface.co/TechWolf/JobBERT-v2}
\end{enumerate}

By combining these contributions, we provide a scalable and efficient framework for labor market analysis.
In \cref{sec:related} we will lay out the previous work on both skill extraction and job title normalization. 
The methodologies for \contextmatch, \skillxl\ and \newjobbert\ are described in \cref{sec:skilltransformer,,sec:benchmark,,sec:jobbert} respectively. 
Experimental results are detailed in \cref{sec:experiments}.

\section{Related work} \label{sec:related}

We review prior research on both skill extraction and job title normalization. 
This review focuses on some key limitations of existing approaches, particularly regarding computational efficiency and the trade-off between model complexity and performance\,---\,challenges we address throughout this work.

\subsection{Skill Extraction}

Skill extraction is a foundational task in labor market analysis that identifies and standardizes skill mentions in unstructured job advertisements. 
This task is central to HR applications, such as resume screening, job recommendations, and workforce planning. 
However, skill extraction poses unique challenges owing to the variability of natural language in job ads and the need to handle explicit and implicit skill descriptions.

Early skill extraction methods identified only skill mentions without normalizing them to a common taxonomy~\cite{Zhao_Javed_Jacob_McNair_2015,representationjobskill, sayfullina, zhang-etal-2022-skillspan}.
In their simplest form, these methods rely on named entity recognition (NER), either through rule-based matching~\cite{Zhao_Javed_Jacob_McNair_2015,representationjobskill} or through training recurrent neural networks~\cite{sayfullina}. 
A significant advancement came with SkillSpan~\cite{zhang-etal-2022-skillspan}, which reframed the task as a more flexible span detection problem, with recent leading methods such as NNOSE~\cite{zhang-etal-2024-nnose} and Skill-LLM~\cite{herandi2024skill}.
However, by leaving out the normalization towards a common taxonomy, their application to robust large-scale market analysis is limited.

To address the normalization challenge, subsequent work approached skill extraction as an extreme multi-label classification (XMLC) problem.
These methods map text to predefined skill taxonomies such as ESCO or O*NET~\cite{decorte2022design,gnehm-etal-2022-fine,decortellm,clavieskill,d2024context}, inspired by dense-encoder XMLC methods such as LightXML~\cite{jiang2021lightxml}, DeepXML~\cite{dahiya2021deepxml} and XR-Transformer~\cite{zhang2021fast}.
In our earlier work~\cite{decorte2022design}, we used binary logistic regression classifiers for each ESCO skill, trained on rule-mined data.
We also contributed the SkillSpan-ESCO benchmark, enabling the first systematic evaluation of fine-grained skill label predictions.

Because of the computational challenges of training thousands of binary classifiers in XMLC tasks, contrastive learning has emerged as an efficient alternative for skill extraction.
Contrastive learning builds on the seminal contrastive loss method~\cite{hadsell2006dimensionality}, further popularized by the InfoNCE objective~\cite{oord2018representation}, and large-scale instantiations such as SimCLR~\cite{chen2020simple} and CLIP~\cite{radford2021learning}.
Applied to skill extraction, this approach trains a bi-encoder architecture that learns to embed both text and skill labels in a shared vector space, thereby enabling efficient similarity-based skill ranking.
Its first application to skill extraction \cite{gnehm-etal-2022-fine} used a two-stage pipeline for German job ads: first detecting skill spans, then ranking ESCO skills against these spans, based on a contrastive learning approach.
We proposed a significant simplification in an earlier contribution~\cite{decortellm} by eliminating the span detection step and directly learning to rank skills against complete sentences, thus enabling the capture of both explicit and implicit skill mentions.

Recent advances in skill extraction have been driven by large language models (LLMs) that offer two distinct advantages.
First, their strong language understanding capabilities enable the automatic generation of high-quality training data for skill extraction~\cite{decortellm, clavieskill}, thereby addressing the data scarcity challenge.
Second, LLMs excel at few-shot learning, leading to new extraction approaches that leverage prompting.
For instance,~\cite{clavieskill} demonstrated success with a two-stage approach that first retrieves candidate skills and then uses LLM prompting for final selection.
This approach was further refined by~\cite{d2024context}, which introduced methods to automatically optimize prompts, achieving state-of-the-art results.
However, these LLM-based methods face practical limitations: they incur significant computational costs and latency compared with local encoder-based approaches, making them less suitable for large-scale applications requiring real-time processing.

The contrastive learning method to skill extraction~\cite{decortellm} has the smallest, yet still very considerable performance gap compared to LLM-based systems. 
We hypothesize that this at least partially stems from ranking predictions solely at the sentence level, neglecting valuable token-level information.
By incorporating token-level information into the skill extraction method, we aim to both improve recall and enhance the interpretability of predictions, while remaining much more efficient than LLM-based systems.

\subsection{Job title normalization}

Job title normalization has traditionally been considered as a (semi-)supervised learning problem. 
A first effort by LinkedIn defined a ``couple dozen'' standard job categories and performed classification based on common key phrases~\cite{highprecphrase}.
A more elaborate taxonomy of over 4k job titles was used in \textbf{Carotene}, using a hierarchical cascade model for job title classification~\cite{carotene}. 
The same task was later solved by \textbf{DeepCarotene}, using an end-to-end deep convolutional neural network~\cite{deepcarotene}.
These methods have a disadvantage as they require extensive annotation of training examples for each job title class. 

To overcome this, we introduced \textbf{JobBERT}~\cite{jobbert}, where we pioneered job title normalization as a ranking task, indicating similarity of job titles based on a specialized representation learning method. 
Specifically, we developed a BERT-based job title representation encoder trained on noisy skill requirements from job ads.
This method does not require annotating training data, but instead relies on the assumption that good representations of job titles can be achieved by learning to predict the probability of skill requirements conditioned on the job ad title.
This method was adapted in the \textbf{Doc2VecSkill} work \cite{Zbib2022LearningJT} through separate stages for learning representations for skills and then for job titles.
Later, \textbf{VacancySBERT} \cite{VacancySBERT} simplified the approach by training a siamese network with contrastive learning and in-batch negatives to draw job title representations closer to the representation of concatenated skill sets from job ads, where skills are extracted using a non-disclosed proprietary algorithm.
The most performant approach to job title normalization is the so-called \textbf{Job Description Aggregation Network}, which omits the explit skill data requirement and instead learns good job title representations by learning to match them to representations of corresponding job ad descriptions~\cite{laosaengpha-etal-2024-learning}.

We build on the insights of previous work, bringing together a new method of contrasting job titles and corresponding skill sets, with full transparency of the skill extraction method as also introduced in this work.
Our JobBERT v2 approach is most similar to VacancySBERT but considers the asymmetry of matching job titles to skill sets.

\Figure[t!](topskip=0pt, botskip=0pt, midskip=0pt){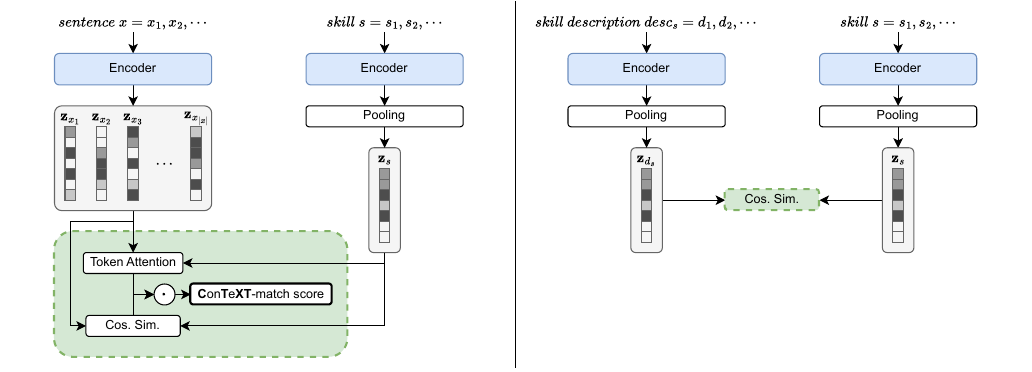}
{ \textbf{Overview of the multi-task setup proposed for our skill extraction training procedure. The bi-encoder produces token embeddings for sentences $x$ and averaged embeddings for skills $s$ and their descriptions $desc_s$. The \contextmatch\ mechanism is applied to produce the matching scores between sentence and skill (left).
For the skill descriptions, simple cosine similarity is used instead of the \contextmatch\ scores (right).}\label{fig:overview}}

\section{Skill Extraction Methodology} \label{sec:skilltransformer}

Our skill extraction approach aims to identify relevant skills in job ad sentences with minimal annotation requirements, following the philosophy of \cite{decortellm}. 
Instead of fully labeled text spans, we require only pairs of a sentence and an associated skill label. 
This relaxed data requirement enables scalable training by avoiding expensive, exhaustive span-level annotations. 
We train a bi-encoder with contrastive learning based on this training data. 
The key innovation that we introduce here is \textbf{Con}trastive \textbf{T}oken-level \textbf{eX}plainable \textbf{T}ext \textbf{match}ing (\contextmatch).
It is an adaptation of contrastive text representation learning in which we remove the information bottleneck of averaging the sentence representations, instead allowing label-dependent token-level attention.
Thus, the model can better attend to the parts of a sentence that are most indicative of a given skill, providing both accurate retrieval and interpretable attribution.

\subsection{Token-level Contrastive Learning}

Our model architecture is a bi-encoder transformer network that outputs contextual representations of both sentences and skills. 
For a given sentence $x$, all skills $s$ are ranked based on a metric $\text{match}(x, s)$. 
Rather than using the cosine similarity of the representations averaged across the tokens, we introduce label-dependent token-level similarity aggregation through \contextmatch. 
This method calculates the cosine similarity between the averaged skill representation and each token representation of the sentence. 
The final match score is defined as a weighted average of the token similarities through a simple multiplicative attention mechanism.

Specifically, given a sentence $x$ and skill $s$, they are tokenized into ${x_1, x_2, \dots, x_{|x|}}$ and ${s_1, s_2, \dots, s_{|s|}}$ respectively.
The bi-encoder computes the contextual token representations for the sentence and the skill independently (through full weight sharing), represented by ${\mathbf{z}_{x_i}}$ and ${\mathbf{z}_{s_i}}$.
We represent the skill $s$ by averaging the embeddings of its tokens:

\begin{equation}
\mathbf{z}_s = \frac{1}{|s|} \sum_{t=1}^{|s|} \mathbf{z}_{s_t}.
\label{skillembedding}
\end{equation}

The attention mechanism operates on the averaged skill embedding $\mathbf{z}_s$ and individual sentence token embeddings $\mathbf{z}_{x_i}$.
The cosine similarity between the skill embedding and token $x_i$ is given by:

\begin{equation}
\text{cos}(x_i, s) = \frac{\mathbf{z}_{x_i} \cdot \mathbf{z}_s}{\|\mathbf{z}_{x_i}\| \cdot \|\mathbf{z}_s\|}.
\label{tokensimilarity}
\end{equation}

Finally, the match between the sentence and the skill is obtained as the weighted average of the token similarities, as follows:

\begin{equation}
\text{match}(x, s) = \sum_{j=1}^{|x|} \alpha_j \cdot \text{cos}(x_j, s),
\label{eq:match}
\end{equation}

\noindent where the weights $\alpha_j$ sum to one, as defined by

\begin{equation}
\alpha_j = \frac{\exp\left(\mathbf{z}_{x_j} \cdot \mathbf{z}_s \right)}{\sum_{k=1}^{|x|} \exp\left(\mathbf{z}_{x_k} \cdot \mathbf{z}_s \right)}.
\label{weights}
\end{equation}

This token-level attention mechanism eliminates the information bottleneck that is otherwise imposed by averaging the sentence representation into a fixed-length embedding, instead allowing skill matching to dynamically attend to the relevant parts of the sentence.
Second, it allows for native attribution and visualization of which parts in the text are matched to a certain skill.
Skills are ranked by relevance with respect to a sentence $x$ by descending $\text{match}\left(x, s\right)$.


Our training method requires a pairwise dataset of sentences and corresponding skills. 
Formally, the training dataset $\mathcal{D}$ consists of pairs $(x, s)$ where $x$ is a sentence from a job ad and $s$ is a corresponding skill label. 
The training procedure is based on contrastive learning with the InfoNCE framework as introduced in~\cite{oord2018representation} relying on in-batch negatives, and adapted to its symmetric variation.  
The loss for a given pair $(x, s)$ in a batch of size $B$ is defined as:

\begin{equation}
\mathcal{L}_{x,s} = \frac{\mathcal{L}_{x,s}^{\text{forward}} + \mathcal{L}_{x,s}^{\text{backward}}}{2},
\label{symmloss}
\end{equation}

where
\begin{equation}
\mathcal{L}_{x,s}^{\text{forward}} = - \log \frac{\exp(\text{match}(x, s) \cdot \text{scale})}{\sum_{k=1}^{B} \exp(\text{match}(x, s_k) \cdot \text{scale})},
\label{forwardloss}
\end{equation}
and
\begin{equation}
\mathcal{L}_{x,s}^{\text{backward}} = - \log \frac{\exp(\text{match}(x, s) \cdot \text{scale})}{\sum_{k=1}^{B} \exp(\text{match}(x_k, s) \cdot \text{scale})}.
\label{backwardloss}
\end{equation}

Here, the \(\text{scale}\) hyperparameter controls the entropy of the softmax function.
We rely on the gradient caching technique from \cite{gao-etal-2021-scaling} to effectively scale up the \contextmatch\ method to very large batch sizes, by splitting them into micro-batches of size 512.

Finally, we make use of rich meta-data that is often present in skill taxonomies. 
We specifically make use of skill descriptions as present in e.g. ESCO, because the assumption is that these descriptions should aid in generalization of the final model. 
The contrastive task of matching skills with their corresponding description makes use of the same symmetrical InfoNCE loss with the exception of using simple cosine similarity between the averaged representations of both instead of the \contextmatch\ scores.
The complete setup is illustrated in~\cref{fig:overview}.
During training, batches are always constructed from pairs of one task only, and the tasks are sampled proportionally to their respective training data set sizes.
We note that more information from skill taxonomies can be incorporated into this multi-task setup, and we show the impact of some of these choices in Appendix~\ref{app:ablations}.

\subsection{Calibration and redundancy filtering}

The trained model produces a ranked list of skill predictions, which requires the calibration of a threshold $\tau$ that we select for maximal F1 score.
In addition to retaining relevant skills through a calibrated threshold, we are also concerned about the \textit{redundancy} of the returned labels, as large label sets are often not mutually exclusive, and excessive redundancy in the returned labels can degrade user experience. 
An example of such redundant skills is ``machine learning'' and ``utilise machine learning'' which are two separate skills present in ESCO.
To handle semantic overlap among the predicted skills, we introduce a redundancy filtering step using the \contextmatch's token-level attention mechanism.
We assume that when two semantically overlapping labels are matched to a sentence, their corresponding token-level attention scores should follow a similar pattern. 
From the skills that meet threshold $\tau$, we retain only those that have the highest dot product with at least one token in the input sequence. 
The dot product between the skill representation and the token representation is used instead of the attention scores $\alpha$, as we empirically find them to work best. 
This may be because of the normalizing effect of the softmax function, which foregoes the comparability of the scores across multiple labels. 
The template tokens of the model (\textit{beginning of sentence} and \textit{end of sentence}) are not considered in this selection mechanism.

\section{Skill-XL Benchmark Development} \label{sec:benchmark}

We refer to \cite{senger-etal-2024-deep} for an in-depth survey of the existing benchmarks for skill extraction. 
Most benchmarks are tailored to span labeling without linking the mentions to a standardized fine-grained classification of skills. 
There are two exceptions: \cite{bhola-etal-2020-retrieving} retrieves job advertisements from a government platform\footnote{https://www.mycareersfuture.gov.sg/} where the employers also added fine-grained skill labels.
The labels are not linked to the relevant sentences in the job ad, making them unsuitable for evaluation at the sentence level. 
Second, in our previous work~\cite{decorte2022design}, we published fine-grained ESCO skill labels linked on top of the span labeling benchmark ``SkillSpan'' by \cite{zhang-etal-2022-skillspan}.
While this benchmark, referred to as \textbf{SkillSpan-ESCO}, allows sentence-level evaluation, it does not take into account the redundancy of the large number of skills in ESCO.
In other words, the SkillSpan-ESCO data has been annotated with just one representative ESCO skill per skill in the sentence, even when multiple correct ESCO skills apply. 
This method of annotation leads to an imperfect evaluation of skill extraction.

Based on these insights, we set out to develop \skillxl\ as the first benchmark for skill extraction that has both fine-grained and exhaustive ESCO skill annotation on the sentence level, with equally informative labels for a sentence being clustered into groups.
We perform the annotation for full job ads such that \skillxl\ can also be used for document-level evaluation.

A team of 12 artificial intelligence experts annotated up to 12 full job ads each, all sampled from the TechWolf data lake, posted in the United States in January 2024.
Because the distribution of roles for which job ads are written is skewed, we sample ads for annotation in two different ways.
Each expert received a set of five randomly sampled job ads, as well as five \textit{diverse} job ads that were better spread out over the support of the data distribution. 
The selection of these diverse job ads was based on a proprietary job ad feature representation that was subject to FAst Diversity Subsampling~\cite{Shang2023}.
Finally, each expert was asked to annotate two more job ads which were selected from the other annotators' data, as to allow for measuring inter-annotator agreement.

\Figure[t!]{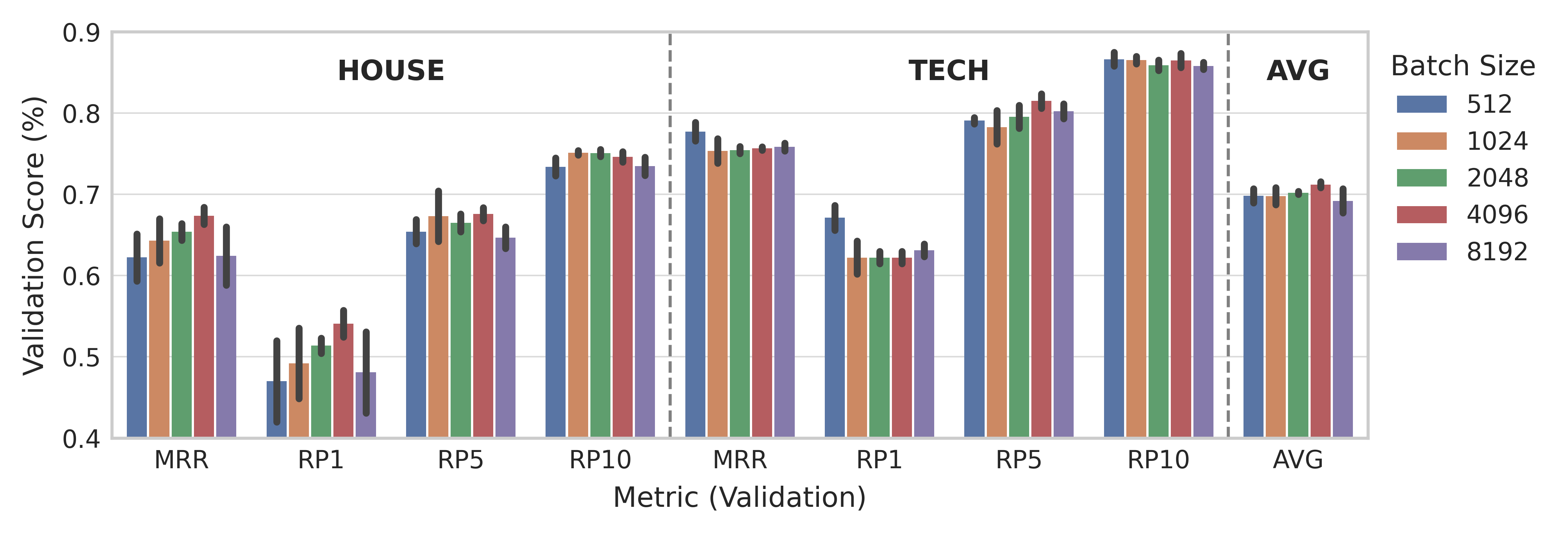}
{\textbf{ Average performance on the SkillSpan–ESCO development set for varying batch sizes. Each bar shows the mean score over three training runs with different random seeds; black whiskers denote one standard deviation.}\label{fig:batchsize}}

The average job ad contained 49 sentences, of which an average of 23 were identified as relevant by a proprietary segmentation model.
Annotation of one job ad took around 30 minutes on average to annotate for one annotator. 
We make use of the clustered annotations to define the relevant metrics for the \skillxl\ benchmark.
The precision of skill extraction is defined as the percentage of predicted skill labels that are present in a reference skill cluster. 
Recall on the other hand is now calculated as the percentage of annotated skill clusters that are represented by at least one label in the predictions. 
Finally, the balanced optimization of precision and recall is expressed as the F1 score and calculated as the harmonic mean of precision and recall.

In order to measure inter-annotator agreement, we used the job ads where two or more annotators provided annotations. 
Conventional agreement metrics such as Cohen's kappa are less suitable for multi-label classification tasks~\cite{marchal-etal-2022-establishing}.
We therefore revert to the averaged value of the F1 scores obtained by in turn evaluating each annotator's labels with another one's labels as reference.
Doing so results in an average F1 agreement score of 0.4395, which we consider high given the large number of labels.
Where multiple experts annotated the same job ad, we select the annotations from the expert with the highest average F1 agreement score as a simple proxy for the quality of their annotations.
Finally, all job ad titles and sentences were anonymized by masking job-related sensitive and personal data regarding organization, location, contact, and name, following~\cite{jensen-etal-2021-de}.

\begin{table}[h]
\centering
\caption{\skillxl\ statistics. The table presents the number of job ads, sentences, and related statistics across different splits.}
\begin{tabular}{c|lrr||>{\columncolor[gray]{0.9}}r}
\toprule
& Statistics & \textbf{RANDOM} & \textbf{UNIQUE} & \textbf{TOTAL} \\ \midrule
\multirow{5}{*}{\rotatebox[origin=c]{90}{\textbf{Development}}} 
    & \# Job ads            & 28    & 27    & 55     \\
    & \# Sentences          & 1,650 & 1,039 & 2,689   \\
    & \# Relevant sentences & 894   & 435 & 1,329 \\
    & \# Skill clusters     & 1,055   & 721   & 1,776  \\
    & \# Skills             & 2,262 & 1,695   & 3,957  \\
\midrule
\multirow{5}{*}{\rotatebox[origin=c]{90}{\textbf{Test}}} 
    & \# Job ads            & 28 & 28 & 56 \\ 
    & \# Sentences          & 1,428 & 1,298 & 2,726 \\ 
    & \# Relevant sentences & 728 & 512 & 1,240 \\ 
    & \# Skill clusters     & 1,103 & 795 & 1,898 \\
    & \# Skills             & 2,685 & 1,829 & 4,514 \\ 
\bottomrule
\end{tabular}
\label{tab:dataset_stats}
\end{table}

The resulting \skillxl\ statistics are shown in~\cref{tab:dataset_stats}. 
The benchmark provides a valuable resource for evaluating skill extraction models, with its granularity and diversity making it broadly applicable to a range of methods, including span-based and sequence-labeling approaches. An annotated job ad example is shown in Appendix~\ref{app:example}.

\section{Skill-based Job Title Normalization} \label{sec:jobbert}

For job title normalization, we follow the idea of using job-skill information to learn strong job title representations\,---\,as originally proposed in JobBERT~\cite{jobbert}.
Compared to the original study that used literal occurrence to find ESCO skills, we use qualitative skills extracted with our trained skill extraction model. 
We decide to use a simple contrastive learning strategy with the InfoNCE loss, optimizing a transformer encoder architecture to distinguish the corresponding skill set for a job title from the other skill sets in the batch. 
Rather than using complete weight-sharing as is common with sentence transformer models, we add an asymmetrical linear layer which accommodates for the different semantic meaning of job titles on the one hand and skill sets on the other.

\section{Experiments and results} \label{sec:experiments}

\subsection{Skill Extraction}

For the skill extraction model, we used ESCO (v1.1.0) as the taxonomy of choice because of its prominent position in skill extraction research and its rich meta-data.
The taxonomy contains 13,981 unique skills, each with a description.
For the job ad sentence training data, we used the synthetic dataset from~\cite{decortellm} which contains up to 10 synthetically generated job ad sentences for each skill in ESCO, totalling 138,260 unique sentence-skill pairs.
For evaluation, we used both \textbf{SkillSpan-ESCO} to compare to the performance of previous methods, as well as \skillxl.
The latter is particularly used for the calibration of the final model, as well as for measuring the prediction redundancy.

For both \textbf{SkillSpan-ESCO} and \skillxl\, we used the ranking-based evaluation metric macro-averaged R-Precision@K (RP@K)~\cite{chalkidis-etal-2019-extreme}.
Because predictions are made on a sentence-basis, we restrict the evaluation to low values of K. 
RP@K is defined in \eqref{eq:RP}, where the quantity $\text{Rel}(n,k)$ is a binary indicator of whether the $k^\text{th}$ ranked label is a correct label for data sample $n$, and $R_n$ is the number of gold labels for sample $n$.
In addition, we use the mean reciprocal rank (MRR) of the highest ranked correct label as an indicator of the ranking quality.

\begin{equation}
    \text{RP@K} = \frac{1}{N} \sum\limits_{n=1}^{N} \sum\limits_{k=1}^{K} \frac{\text{Rel}\left(n, k\right)}{\min\left(K, R_n\right)}
    \label{eq:RP}
\end{equation}

We trained each model for a maximum of one epoch, and build in an early-stopping criterion that measures RP@5 on the SkillSpan-ESCO development set every 10\% of the epoch and halts when there has been no increase two consecutive times.
A sentence embedding model based on MPNET~\cite{mpnet} and further trained on 1B positive pairs was used as initialization.\footnote{https://huggingface.co/sentence-transformers/all-mpnet-base-v2}
The model contains 109 million parameters and has a default maximum sequence length of 384 tokens which we keep. 
As our method predicts skills sentence by sentence, this sequence length is sufficiently long for practical purposes.
The maximum sentence length observed in the TECH and HOUSE validation sets is 178 tokens, with a median of 16 tokens.
The AdamW optimizer is used with WarmupLinear learning rate scheme, learning rate $5\mathrm{e}{-5}$ and 10\% of the data for the warmup window.
The $\text{scale}$ hyperparameter is kept at its default value of 20. 

\begin{table*}[]
\centering
\caption{Performance comparison of skill extraction methods across the SkillSpan-ESCO test sets (HOUSE, TECH, TECHWOLF). Metrics include Mean Reciprocal Rank (MRR) and Recall Precision at Top-K (RP@1, RP@5, RP@10). Following previous work, the RP@K scores are reported in percentage points. Average positive training examples per skill are reported in the first column (N). Results of previous methods are the reported scores, and empty cells mean the metric was not reported. The strongest results per metric are shown in bold, and the second strongest are underlined. Whenever our results are significantly ($p < 0.05$) stronger than the second-best method, we indicate this with *.}
\label{tab:performance}
\addtolength{\tabcolsep}{-0.2em}
\begin{tabular}{@{}l|c|cccc|cccc|cccc@{}}
\toprule
& & \multicolumn{4}{c|}{\textbf{HOUSE}} & \multicolumn{4}{c|}{\textbf{TECH}}  & \multicolumn{4}{c}{\textbf{TECHWOLF}} \\\midrule
 & \textbf{N} & \textbf{MRR} & \textbf{RP@1} & \textbf{RP@5} & \textbf{RP@10} & \textbf{MRR} & \textbf{RP@1} & \textbf{RP@5} & \textbf{RP@10} & \textbf{MRR} & \textbf{RP@1} & \textbf{RP@5} & \textbf{RP@10}     \\\midrule
\rowcolor{gray!25} \multicolumn{14}{c}{\textbf{Encoder classifiers}} \\ \midrule
Decorte \textit{et al.} \cite{decorte2022design} & 365 & 0.299 &  -- & 30.82 & 38.69 & 0.339 & -- & 31.71 & 39.19 & -- & -- & -- & -- \\
Clavi\'e \textit{et al.} \cite{clavieskill} & 40 & 0.326          & 27.20          & 37.60          & 46.47          & 0.299         & 27.16          & 33.41          & 39.86          &  --           & --             &   --           &   --           \\ \midrule
\rowcolor{gray!25} \multicolumn{14}{c}{\textbf{Encoder rankers}} \\ \midrule
Clavi\'e \textit{et al.} \cite{clavieskill} & 40 & 0.355          & 26.44          & 35.22          & 43.73          & 0.405         & 32.84          & 49.67          & 58.66          &  --           &  --            &  --            &   --           \\
Decorte \textit{et al.} \cite{decortellm} & \textbf{10} & 0.426    & 27.10          &  45.94          & 53.87          & 0.521         & 38.46   & 54.19          & 61.52          & 0.506   & 37.42    & 52.64    & 60.10    \\
\contextmatch\ (Ours)  & \textbf{10} & \textbf{0.530}*     & \underline{38.42}          & 51.09          & \underline{65.84}    & \textbf{0.632}* & \textbf{50.99}*    & \textbf{63.98}    & \textbf{73.99}*    & \textbf{0.562}* & \textbf{43.15}* & \textbf{57.69}* & \textbf{66.08} \\\midrule
\rowcolor{gray!25} \multicolumn{14}{c}{\textbf{LLM-based systems}} \\ \midrule
Clavi\'e \textit{et al.} \cite{clavieskill} \tiny{GPT3.5} & 40 & 0.427 & 36.92 & 43.57 & 51.44 & 0.488 & 40.53 & 52.50 & 59.75 & -- & -- & -- & -- \\
Clavi\'e \textit{et al.} \cite{clavieskill} \tiny{GPT4}   & 40 & \underline{0.495} & \textbf{40.70} & \underline{53.34} & 61.02 & \underline{0.537} & \underline{46.52} & \underline{61.50} & 68.94 & -- & -- & -- & \\
IReRa \cite{d2024context} & n/a & -- & -- & \textbf{56.50} & \textbf{66.51} & -- & -- & 59.61 & \underline{70.23} & -- & -- & \underline{57.04} & \underline{65.17} \\\bottomrule
\end{tabular}
\end{table*}

We use the augmentation trick that we introduced in \cite{decortellm}, which is to either prepend or append (equally split) each job ad sentence with another random job ad sentence during training.
The idea behind this augmentation strategy is to overcome the fact that the synthetic job ad sentences tend to focus on just one skill, whereas we want the encoder to be able to effectively model multiple topics in its input.
Note that this augmentation technique is only applied to the job ad sentences, not to skills or descriptions.

Contrastive learning with in-batch negatives has generally been shown to benefit significantly from large batch sizes~\cite{SimCLR}.
Therefore, we decided to focus on determining the optimal batch size first, while keeping all other hyper parameters untouched. 
\cref{fig:batchsize} shows the results obtained for the SkillSpan-ESCO development set for different batch sizes. 
For each batch size, we trained three models with different random seeds, and the standard deviation is shown as error bars. 
Based on this analysis, we selected batch size 4,096 as optimal, given that it obtains the highest average metric values, with lower standard deviation than observed for smaller batch sizes.
For further analysis of training decisions, we refer to Appendix~\ref{app:ablations}.

The final model results on the test set of SkillSpan-ESCO are shown in~\cref{tab:performance}. 
A detailed runtime and cost comparison with the recent IReRa pipeline is provided in Appendix~\ref{app:comparison}.
The results demonstrate that \contextmatch\ outperforms all other encoder-based methods, and even outperforms the LLM-based systems on most metrics.

Finally, we calibrate the model predictions using the total \skillxl\ development set.
The optimal $\tau$ was searched between 0 and 1, with steps of 0.01. The value $\tau=0.53$ yields the highest F1 score of 0.4100, with an average redundancy of 27.32\%. 
We formally define the redundancy of a prediction as the highest fraction of true positives that can be left out while still having the same clusters represented by at least one label in the predictions.
When applying the intelligent filtering, in which the predictions are filtered based on their token-level attention, the optimal threshold was found to be 0.48, and an F1 score of 0.4389 was reached with a significantly reduced redundancy of only 13.46\%.

The calibrated model achieves an F1 score of 0.407 and 0.401 on the \skillxl\ RANDOM and UNIQUE test sets respectively, with a corresponding prediction redundancy of 16.88\% and 15.38\%.
These numbers serve as the first baseline results for our new \skillxl\ benchmark.

\subsection{Job title normalization}

To train \newjobbert, we start from the same base sentence embedding model (based on MPNET~\cite{mpnet}) as we did for the skill extraction model.\footnote{https://huggingface.co/sentence-transformers/all-mpnet-base-v2}
We instantiate the asymmetrical linear layer to project the original 768-dimensional representations to a 1,024-dimensional space.
To train \newjobbert, we first enriched a large number of job ads with ESCO skills using the final skill extraction model.
We randomly selected 100,000 job ads per month (from January 2020 to December 2024) posted in the United States. 
Note that we again applied the proprietary algorithm to filter out irrelevant sentences. 
After removing those ads that had fewer than five unique skills tagged, we ended up with a dataset of 5,579,240 enriched job ads, each consisting of a job ad title and the extracted set of skills.
The skills of a job ad are shuffled and combined into one comma-separated text.
Because of the limited context window of the model (512 tokens), we decided to limit the number of skills in a job ad to 25, which we achieve through random selection when necessary.
A resulting pair in the training dataset may then appear as follows:

\begin{itemize}
    \item Title: \textit{Oracle Apps CRM Technical Consultant}
    \item Skills: \textit{think analytically, Oracle Data Integrator, data models, computer technology, customer relationship management, use ICT systems}
\end{itemize}

\begin{table}[H]
\centering
\caption{Performance comparison on the original JobBERT job title normalization benchmark. Metrics include Mean Reciprocal Rank (MRR) and Recall at Top-K (RP@5, RP@10). Results of previous methods are the reported scores. The recall is reported in percentage points.}
\label{tab:titleperformance}
\begin{tabular}{lccc}
\toprule
\textbf{Model}                          & \textbf{MRR}   & \textbf{R@5}   & \textbf{R@10}  \\ 
\midrule
JobBERT V1~\cite{jobbert}               & 0.309          & 38.65          & 46.04          \\
Doc2VecSkill~\cite{Zbib2022LearningJT}  & 0.341          & 45.95          & 54.00          \\
JD Aggregation Network~\cite{laosaengpha-etal-2024-learning} & 0.387          & 49.24          & 57.22          \\ 
\midrule
\textbf{\newjobbert\ (Ours)}            & \textbf{0.390} & \textbf{50.08} & \textbf{58.47} \\
\bottomrule
\end{tabular}
\end{table}

We trained the model for one full epoch, with the InfoNCE loss function, batch size of 2\,048, scale of 20 and learning rate $5\mathrm{e}{-5}$, and linear learning rate decay without warm-up.
The resulting performance is shown in~\cref{tab:titleperformance}, from which can be seen that the model outperforms all previous baselines even with minimal hyperparameter tuning. 
This demonstrates the power of the qualitative (skills) training data and the simple yet effective training objective we proposed or \newjobbert.

\subsection{Visualization of Skill Extraction}

We can visualize the token-level attention scores of \contextmatch\ to explain based on what tokens a skill was predicted.
Three sentences, alongside their predicted skill tags and respective token-level explanation are shown in~\cref{tab:tokenvisual}.
An alignment study between these token-level explanations and human rationales is reported in Appendix~\ref{app:interpretability}.

\addtolength{\tabcolsep}{-0.4em}

\begin{table}[H]

\renewcommand{\arraystretch}{1.5} 
\centering

\caption{Token attention scores for each skill predicted, for three different example sentences. Higher attention scores are visualized as a stronger background color for the corresponding token.}
\label{tab:tokenvisual}

\begin{tabular}{|c|c|c|c|c|c|c|c|c|c|c|} \hline
\rowcolor{gray!25} \multicolumn{11}{|c|}{\textbf{(a) Lead the group in charge of cost and risk management objectives}} \\ \hline
\rowcolor{gray!10} \multicolumn{11}{|c|}{\textbf{cost management}} \\
\cellcolor{red!1.07}lead & \cellcolor{red!2.30}the & \cellcolor{red!3.34}group & \cellcolor{red!0.85}in & \cellcolor{red!2.56}charge & \cellcolor{red!5.38}of & \cellcolor{red!59.12}cost & \cellcolor{red!4.27}and & \cellcolor{red!5.76}risk & \cellcolor{red!9.45}management & \cellcolor{red!4.22}objectives \\ \hline
\rowcolor{gray!10} \multicolumn{11}{|c|}{\textbf{lead a team}} \\
\cellcolor{red!54.18}lead & \cellcolor{red!9.00}the & \cellcolor{red!10.80}group & \cellcolor{red!6.27}in & \cellcolor{red!12.09}charge & \cellcolor{red!0.88}of & \cellcolor{red!0.94}cost & \cellcolor{red!0.48}and & \cellcolor{red!0.55}risk & \cellcolor{red!0.75}management & \cellcolor{red!0.62}objectives \\ \hline
\rowcolor{gray!10} \multicolumn{11}{|c|}{\textbf{risk management}} \\
\cellcolor{red!1.26}lead & \cellcolor{red!2.36}the & \cellcolor{red!1.64}group & \cellcolor{red!1.83}in & \cellcolor{red!1.76}charge & \cellcolor{red!5.21}of & \cellcolor{red!10.63}cost & \cellcolor{red!9.44}and & \cellcolor{red!42.54}risk & \cellcolor{red!14.25}management & \cellcolor{red!6.36}objectives \\ \hline
\end{tabular}  \\[1em]

\begin{tabular}{|c|c|c|c|c|c|c|c|c|c|c|c|} \hline
\rowcolor{gray!25} \multicolumn{12}{|c|}{\textbf{(b) You will write software in Java, Python and C++}} \\ \hline
\rowcolor{gray!15} \multicolumn{12}{|c|}{\textbf{C++}} \\
\cellcolor{red!0.68}you & \cellcolor{red!1.24}will & \cellcolor{red!1.05}write & \cellcolor{red!2.58}software & \cellcolor{red!2.38}in & \cellcolor{red!3.74}java & \cellcolor{red!1.63}, & \cellcolor{red!2.49}python & \cellcolor{red!1.59}and & \cellcolor{red!19.17}c & \cellcolor{red!31.57}+ & \cellcolor{red!29.13}+ \\  \hline
\rowcolor{gray!20} \multicolumn{12}{|c|}{\textbf{authoring software}} \\
\cellcolor{red!7.64}you & \cellcolor{red!6.21}will & \cellcolor{red!33.91}write & \cellcolor{red!24.94}software & \cellcolor{red!6.49}in & \cellcolor{red!1.51}java & \cellcolor{red!4.67}, & \cellcolor{red!1.00}python & \cellcolor{red!4.59}and & \cellcolor{red!0.32}c & \cellcolor{red!0.47}+ & \cellcolor{red!0.86}+ \\  \hline
\rowcolor{gray!15} \multicolumn{12}{|c|}{\textbf{Java (computer programming)}} \\
\cellcolor{red!1.91}you & \cellcolor{red!3.01}will & \cellcolor{red!3.55}write & \cellcolor{red!4.51}software & \cellcolor{red!8.44}in & \cellcolor{red!45.37}java & \cellcolor{red!13.59}, & \cellcolor{red!3.18}python & \cellcolor{red!4.98}and & \cellcolor{red!2.89}c & \cellcolor{red!2.89}+ & \cellcolor{red!2.75}+ \\  \hline
\rowcolor{gray!15} \multicolumn{12}{|c|}{\textbf{Python (computer programming)}} \\
\cellcolor{red!1.88}you & \cellcolor{red!2.62}will & \cellcolor{red!3.32}write & \cellcolor{red!3.57}software & \cellcolor{red!7.70}in & \cellcolor{red!4.43}java & \cellcolor{red!6.64}, & \cellcolor{red!40.87}python & \cellcolor{red!5.29}and & \cellcolor{red!10.91}c & \cellcolor{red!5.33}+ & \cellcolor{red!4.41}+ \\  \hline
\end{tabular} \\[1em]

\begin{tabular}{|c|c|c|c|c|c|c|c|c|c|c|} \hline
\rowcolor{gray!25} \multicolumn{11}{|c|}{\textbf{(c) Responsible for diagnosing, repairing, and maintaining cars}} \\ \hline
\rowcolor{gray!10} \multicolumn{11}{|c|}{\textbf{diagnose problems with vehicles}} \\
\cellcolor{red!2.62}responsible & \cellcolor{red!3.55}for & \cellcolor{red!23.84}dia & \cellcolor{red!27.78}gno & \cellcolor{red!9.97}sing & \cellcolor{red!4.06}, & \cellcolor{red!3.49}repairing & \cellcolor{red!4.26}, & \cellcolor{red!3.40}and & \cellcolor{red!3.18}maintaining & \cellcolor{red!11.85}cars \\ \hline
\rowcolor{gray!10} \multicolumn{11}{|c|}{\textbf{carry out repair of vehicles}} \\
\cellcolor{red!4.28}responsible & \cellcolor{red!4.83}for & \cellcolor{red!3.55}dia & \cellcolor{red!5.36}gno & \cellcolor{red!9.70}sing & \cellcolor{red!9.05}, & \cellcolor{red!14.35}repairing & \cellcolor{red!6.64}, & \cellcolor{red!10.18}and & \cellcolor{red!12.87}maintaining & \cellcolor{red!16.08}cars \\ \hline
\rowcolor{gray!10} \multicolumn{11}{|c|}{\textbf{maintain vehicle service}} \\
\cellcolor{red!3.98}responsible & \cellcolor{red!4.08}for & \cellcolor{red!2.56}dia & \cellcolor{red!3.54}gno & \cellcolor{red!7.33}sing & \cellcolor{red!8.37}, & \cellcolor{red!9.77}repairing & \cellcolor{red!8.16}, & \cellcolor{red!12.80}and & \cellcolor{red!24.88}maintaining & \cellcolor{red!12.18}cars \\ \hline
\end{tabular}

\end{table}

\section{Conclusion}

In this work, we have introduced a scalable and efficient approach to labor market analysis by tackling two of its main tasks being skill extraction and job title normalization. 
We proposed \contextmatch, a method that enhances accuracy and explainability of skill extraction, achieving state-of-the-art results, outperforming LLM-based systems at a much reduced computational cost.
While this work focuses on skill extraction, \contextmatch\ is a general method that can be applied to other extreme multi-label classification tasks, offering potential gains in performance across various domains.

Secondly, we train \newjobbert, our second version of JobBERT which also achieves the best performance for the job title normalization task using a simple yet effective training method, made possible by leveraging large-scale high-quality skill data generated by our skill extraction model.

Finally, we have developed and released \skillxl, our comprehensive skill extraction benchmark with sentence-level annotations that explicitly address redundancy in large label spaces.
This benchmark serves as a new evaluation framework that allows for measuring both the accuracy and usefulness of skill extraction methods.

The efficiency of our skill extraction and job title normalization model unlocks the potential for large-scale, real-time labor market analysis in practical applications, which is something that can be further explored in future research. 
Extending the methods and results to non-English languages remains an important topic for future work.
By open-sourcing both our models and the new benchmark, we aim to enable further research and innovation in this field.

\appendices

\section{\break Skill Extraction ablations}\label{app:ablations}

Different adaptations to the skill extraction training mechanism were compared against the final model. 
Their impact is assessed on both development sets of SkillSpan-ESCO in~\cref{tab:interventions}, and we only report RP\@5 for conciseness, selected for its lowest observed standard deviations.

\begin{table}[h!]
\arrayrulecolor{black} 
\centering
\caption{Performance comparison of ablation interventions. Significantly worse or better results are indicated with $\downarrow$ or $\uparrow$ (p < 0.05). }
\label{tab:interventions}
\begin{tabular}{@{}l|cccc@{}}
\toprule
\textbf{Intervention}         & \textbf{House RP@5} &        & \textbf{Tech RP@5}  & \\
\midrule
-                        & $67.568 \pm 0.779$  &             & $81.489 \pm 0.882$    &      \\ \midrule
\rowcolor{gray!15} \multicolumn{5}{c}{\textbf{Training objective}} \\
(a) No ConTeXT           & $58.989 \pm 1.352$  & $\downarrow$  & $74.341 \pm 0.105$    & $\downarrow$     \\
(b) No augmentation              & $58.097 \pm 1.375$  & $\downarrow$  & $77.044 \pm 0.588$    & $\downarrow$     \\
(c) Asymm. loss          & $65.638 \pm 0.284$  & $\downarrow$  & $81.637 \pm 1.008$    &      \\
\rowcolor{gray!15} \multicolumn{5}{c}{\textbf{Training tasks}} \\
(d) No descriptions      & $66.658 \pm 1.057$  &             & $80.896 \pm 1.557$    &      \\
(e) Add synonyms         & $67.960 \pm 1.232$  &             & $75.600 \pm 0.839$    & $\downarrow$     \\
\bottomrule
\end{tabular}
\end{table}

In intervention (a), we replace the \contextmatch\ mechanism with a cosine similarity score between averaged skill and sentence representations, as used in~\cite{decortellm}.
The large drop in performance signifies the strength of \contextmatch.
Intervention (b) leaves out the prepend/append augmentation during training, which also leads to a considerable drop in performance. 
In intervention (c), the symmetrical loss $\mathcal{L}_{x,s}$ is replaced by only the forward loss $\mathcal{L}_{x,s}^{\text{forward}}$.
While this incurs a significant loss on the HOUSE development set, the performance on the TECH set remains comparable.

\begin{table*}[h]
\centering 
\caption{Annotated Example Job Ad. Relevant sentences are annotated with skill clusters and specific skills. Non-relevant sentences are grayed out.}
\begin{tabular}{p{8cm}p{4cm}}
\toprule
\rowcolor{white} \textbf{Sentence} & \textbf{Skills} \\ \midrule
\rowcolor[gray]{0.9} Job Title: & - \\ \cmidrule{1-2}
\multirow{1}{8cm}[-1em]{Bigdata Architect (Data Modeling / Data Architect)} 
& data models; create data models; design database scheme \\ \cmidrule{1-2}
\rowcolor[gray]{0.9} Location: & - \\ \cmidrule{1-2}
\rowcolor[gray]{0.9} location (Onsite) & - \\ \cmidrule{1-2}
\multirow{5}{8cm}[-1em]{Expertise in Hive, Big Data environments (Hadoop preferred), HBase, Spark (Scala preferred but python is also ok)} 
& analyse big data; data mining \\ \cmidrule{2-2}
& Hadoop \\ \cmidrule{2-2}
& SPARK \\ \cmidrule{2-2}
& Scala \\ \cmidrule{2-2}
& Python (computer programming) \\ \cmidrule{1-2}
\multirow{1}{8cm}[-1em]{Should have experience in scaling an application} 
& distributed computing; decentralized application frameworks \\ \cmidrule{1-2}
\multirow{1}{8cm}[-1em]{Extensive experience in Data Modeling} 
& data models; create data models; design database scheme \\ \cmidrule{1-2}
\multirow{1}{8cm}[-1em]{Extensive experience in Data Architect} 
& information architecture; manage ICT data architecture \\ \cmidrule{1-2}
\multirow{1}{8cm}[-1em]{Should have strong Performance (compute and I/O) background} 
& ICT performance analysis methods; advise on efficiency improvements \\ \cmidrule{1-2}
\multirow{2}{8cm}[-1em]{Should have experience in UNIX shell scripting, Jenkins configurations} 
& use scripting programming \\ \cmidrule{2-2}
& Jenkins (tools for software configuration management) \\ \cmidrule{1-2}
\multirow{2}{8cm}[-1em]{Perform Code Reviews and understand the stack} 
& conduct ICT code review \\ \cmidrule{2-2}
& provide technical expertise; consult technical resources \\ \cmidrule{1-2}
\multirow{1}{8cm}[-1em]{Follow best practices while coding} 
& implement ICT coding conventions; ensure adherence to organisational ICT standards \\ \cmidrule{1-2}
\multirow{2}{8cm}[-1em]{Good Knowledge in Machine Learning and Data Science algorithms} 
& machine learning; utilise machine learning; ML (computer programming) \\ \cmidrule{2-2}
& algorithms; data analytics \\ \cmidrule{1-2}
\rowcolor[gray]{0.9} Thanks \& Regards, & - \\ \cmidrule{1-2}
name & - \\ \cmidrule{1-2}
Senior Executive & - \\ \cmidrule{1-2}
\rowcolor[gray]{0.9} - International Consulting & - \\ \cmidrule{1-2}
\rowcolor[gray]{0.9} organization & - \\
\bottomrule
\end{tabular}
\label{tab:annotated_job_ad}
\end{table*}

We also perform some analysis on the influence of the additional training task. 
Specifically, we drop the description-skill matching task in intervention (d) and observe a moderate drop in performance.
Finally, we consider using the skill synonyms provided in ESCO to create a third contrastive training task of skill synonym matching.
This third task uses the same InfoNCE objective as the description-skill matching task. 
As seen in intervention (e), this incurs a significant drop in performance on the TECH set and a small but insignificant performance increase on the HOUSE set.
The TECH set contains mostly technical jobs, leading to more hard skills (like specific programming languages) and less soft skills~\cite{zhang-etal-2022-skillspan}.
Inspection of the ESCO synonyms provided for hard skills reveals some inaccuracies like listing ``Live Script'' as a synonym for ``JavaScript'', which is related but synonymous.
These inaccuracies are a likely cause for the performance drop on the TECH set.

\section{\break Skill-XL example}\label{app:example}

\Cref{tab:annotated_job_ad} contains part of an annotated job ad from the \skillxl\ development set. Each sentence is shown in a separate row, and skills in the same annotation cluster are separated by a semicolon.

\section{\break \contextmatch\ comparison with IReRa}\label{app:comparison}

To substantiate our claim that ConTeXT‑match delivers state‑of‑the‑art skill extraction quality at a fraction of the cost of LLM‑based systems, we compared it with \textbf{IReRa}~\cite{d2024context}, a three‑stage pipeline that combines a local LLaMA‑2‑13B model\footnote{https://huggingface.co/meta-llama/Llama-2-13b-chat-hf}, a dense retriever and a re-ranking step using OpenAI's GPT‑4 accessed through API.
We create predictions on the combined HOUSE and TECH validation sets (136 unique sentences) and compare both the efficiency of the process and the quality of the predictions.

\subsection{Efficiency comparison}

Both systems were deployed on Google Cloud Platform (GCP) using on‑demand resources.\footnote{https://cloud.google.com/compute/all-pricing}
The Llama model uses float16 weights (stored on 2 bytes), so it requires a minimum of 26 GB VRAM to use locally.
\textbf{IReRa} was run on an \texttt{a2‑highgpu‑1g} virtual machine (1$\times$A100‑40\,GB, 12 vCPU, 85 GB RAM) billed at \$4.09\ per hour, plus GPT‑4 API usage at April 2025 prices (\$10 per million input tokens, \$30 per million output tokens).
In contrast, \textbf{\contextmatch} was evaluated on an inexpensive \texttt{e2‑medium} virtual machine (2 vCPU, 4 GB RAM) billed at \$0.10\ per hour.
All virtual machine prices include a 500GB boot disk and are based on April 2025 pricing.

Both systems processed the combined 136 validation sentences of the HOUSE and TECH benchmarks.
\contextmatch\ ran with a batch size of 8, averaging 1.1 seconds per batch -- about 7 sentences per second. The entire run finished in 19 seconds and cost just \$0.00053. The maximum memory usage ran up to 2.2 GB.
IReRa completed in 14 minutes and 44 seconds, or roughly 6.5 seconds per sentence, which comes down to \$1.00 for compute time. Because the current IReRa implementation does not support batching, the GPU usage is not optimized, so the \$1.00 is an upper bound, and the throughput a lowerbound. OpenAI usage added \$3.06, totaling \$4.06 for the IReRa processing.
These experiments show a cost reduction with a factor of 1/7660 when using \contextmatch\ compared to IReRa, as well as strongly reduced compute and memory requirements.
These results confirm that ConTeXT‑match delivers state‑of‑the‑art skill‑extraction quality \emph{without the prohibitive costs and specialised infrastructure} associated with modern LLM pipelines, making large‑scale, affordable deployment feasible.

\subsection{Qualitative comparison}
\label{sec:qualitative}

To understand \emph{how} both systems differ, we manually inspected their outputs for the 136 validation sentences and grouped the discrepancies into four recurrent patterns. Throughout the discussion, predictions from \contextmatch\ are shown in \textcolor{blue}{blue}, while those from IReRa are shown in \textcolor{violet}{violet}. For IReRa we only consider the top--5 ranked labels, mirroring a realistic downstream cut-off.

\paragraph{Precision versus recall}%

\contextmatch\ exhibits a deliberate high-precision bias: it almost never proposes a skill that is not textually or semantically supported.  The flip side is occasional under-extraction.  
For the sentence \emph{“Work on a mix of front-end, back-end and cloud technologies.”} the model yields \textcolor{blue}{\{cloud technologies\}}, omitting the equally explicit \textit{front-end} and \textit{back-end}.  
IReRa, in contrast, returns \textcolor{violet}{\{cloud technologies, implement front-end website design, design cloud architecture, manage cloud data and storage, cloud security and compliance\}}.  
It has a higher coverage, but three of the five predictions (`design cloud architecture’, `manage cloud data and storage’, `cloud security and compliance’) have no lexical anchor in the sentence and would count as false positives.

\paragraph{Lexical stability versus synonym proliferation}%

\contextmatch’s redundancy-filtering outputs a single canonical form per skill, making the list compact and deduplicated across the corpus.  IReRa frequently emits several near-identical variants, inflating the candidate set.  
For a simple requirement such as \emph{“Fluent in written and spoken English”}, \contextmatch\ returns \textcolor{blue}{\{English\}}, whereas IReRa offers all of \textcolor{violet}{\{English, write English, understand written English, understand spoken English, communication\}}.  
While this boosts recall for loosely coupled downstream ontologies, it pushes extra work to any consumer that must cluster the five labels back to one concept.

\paragraph{Hallucinated abstractions versus literal tokens}%

Driven by GPT-4 re-ranking, IReRa is prone to infer abstract or industry-specific abilities that are only tangentially related to the source sentence.  
From the prompt \emph{“Are you ready to work in a dynamic and international team?”} it proposes \textcolor{violet}{collaborate on international energy projects} and \textcolor{violet}{develop international cooperation strategies}.
Conversely, \contextmatch\ stays close to surface evidence, yielding the literal \textcolor{blue}{work in an international environment}.  
While hallucinated abstractions may reveal latent needs, they harm precision in settings where textual faithfulness is mandatory.

\paragraph{Robustness to noisy tokens versus token artefacts}%

\contextmatch\ occasionally seems less robust in case of an abundance of punctuation, abbreviations or parentheses.  
In the phrase \emph{“HTML and CSS (LESS, SCSS, PostCSS)”} it emits the skill \textcolor{blue}{JSSS}.
IReRa correctly outputs \textcolor{violet}{LESS} and \textcolor{violet}{SCSS}.

\vspace{6pt}
\noindent
\textbf{Summary.}  
Qualitatively, \contextmatch\ fails mainly by omission or minor token glitches, whereas IReRa fails by over-generation and synonym duplication.  
For high-precision use-cases (e.g.,\ automated profile completion) our model’s conservative stance and ten-thousand-fold lower cost are decisive advantages.  
When maximum recall is the overriding goal, IReRa’s expansive, noisier outputs may be preferable, provided that post-hoc clustering and human validation steps are affordable.

\section{\break Interpretability Study: Human Alignment of \contextmatch\ Attention}\label{app:interpretability}

In addition to quantitative accuracy, an important requirement for industrial skill extraction systems is \emph{explainability}. 
Because \contextmatch\ produces token-level attention weights for every predicted skill, we performed a small-scale user study to verify whether those weights correspond to the tokens that humans regard as most diagnostic. 
We drew a random sample of 50 sentences from the development split of \skillxl. 
Two independent annotators from the team were shown the sentence and a corresponding ground-truth skill cluster. 
Annotators highlighted all spans of words they considered an \emph{explanation word}, i.e., a word or phrase that justifies the presence of the given skill. 
The instructions emphasized that noncontiguous selections were allowed and that they should abstain if no token qualified. Each annotator worked in isolation. 
We measured rank correlation between the human binary vector $\mathbf h\!\in\!\{0,1\}^{|x|}$ and the \contextmatch\ attention scores $\boldsymbol{\alpha}\!\in\![0,1]^{|x|}$ using Spearman's~$\rho$, computed per sentence and then averaged.
The template tokens of the model (\textit{beginning of sentence} and \textit{end of sentence}) are not included in this annotation nor correlation measurement. Where an annotator labeled no token, that sentence was skipped for that annotator.
The Spearman rank correlation ($\rho$) between the model's attention and Annotator 1 was 0.56, while for Annotator 2 it was 0.50. The spearman correlation between the binary annotations of both annotators was 0.60. These results indicate that the model's attention correlates positively with both annotators (mean $\rho\!\approx\!0.53$), approaching human-human consistency.

Table~\ref{tab:interact_people_attention} and \ref{tab:business_relationships_attention} visualize the human annotations next to the \contextmatch\ attention scores for two specific examples.
The human annotations in table~\ref{tab:interact_people_attention} show that the skill is attributed to the large majority of words in the sentence. 
Correspondingly, the attention scores are spread out across the tokens.

\begin{table}[h]
\centering
\caption{Comparison of model attention scores ($\alpha$) and human annotation, for the sentence \textit{``Interact with different people of all ages and backgrounds''} and skill \textit{``communicate with elderly groups''}. The attention scores are indicated by the background intensity. Binary human annotations are shown in bold.}
\label{tab:interact_people_attention}
\renewcommand{\arraystretch}{1.2} 
\begin{tabular}{@{}l|l|l@{}}
\toprule
\rowcolor{gray!10} $\boldsymbol{\alpha}$ & Annotator 1 & Annotator 2 \\
\midrule
\cellcolor{red!12.30}interact & \textbf{interact} & \textbf{interact} \\
\cellcolor{red!13.46}with & \textbf{with} & \textbf{with} \\
\cellcolor{red!9.59}different & \textbf{different} & different \\
\cellcolor{red!18.67}people & \textbf{people} & \textbf{people} \\
\cellcolor{red!9.97}of & \textbf{of} & \textbf{of} \\
\cellcolor{red!7.04}all & \textbf{all} & \textbf{all} \\
\cellcolor{red!8.67}ages & \textbf{ages} & \textbf{ages} \\
\cellcolor{red!6.27}and & and & and \\
\cellcolor{red!4.16}backgrounds & backgrounds & backgrounds \\
\bottomrule
\end{tabular}
\renewcommand{\arraystretch}{1.0} 
\end{table}

\newpage

In contrast, table~\ref{tab:business_relationships_attention} shows a case where the skill is clearly explained by one confined part of the sentence. 
The model's attention scores center around the same words, with a particularly sharp signal on the token ``relationships''.

\begin{table}[h]
\centering
\caption{Comparison of model attention scores ($\alpha$) and human annotation, for the sentence \textit{``You will also maintain strong relationships with business partners to help shape requirements and provide both analytical and technical support.''} and skill \textit{``build business relationships''}. The attention scores are indicated by the background intensity. Binary human annotations are shown in bold.}
\label{tab:business_relationships_attention}
\renewcommand{\arraystretch}{1.2} 
\begin{tabular}{@{}l|l|l@{}}
\toprule
\rowcolor{gray!10} $\boldsymbol{\alpha}$ & Annotator 1 & Annotator 2 \\
\midrule
\cellcolor{red!2.42}you & you & you \\
\cellcolor{red!1.92}will & will & will \\
\cellcolor{red!1.53}also & also & also \\
\cellcolor{red!3.29}maintain & \textbf{maintain} & \textbf{maintain} \\
\cellcolor{red!5.81}strong & \textbf{strong} & \textbf{strong} \\
\cellcolor{red!38.63}relationships & \textbf{relationships} & \textbf{relationships} \\
\cellcolor{red!15.05}with & \textbf{with} & \textbf{with} \\
\cellcolor{red!15.26}business & \textbf{business} & \textbf{business} \\
\cellcolor{red!8.30}partners & \textbf{partners} & \textbf{partners} \\
\cellcolor{red!1.31}to & to & to \\
\cellcolor{red!0.93}help & help & help \\
\cellcolor{red!0.45}shape & shape & shape \\
\cellcolor{red!1.09}requirements & requirements & requirements \\
\cellcolor{red!0.25}and & and & and \\
\cellcolor{red!0.33}provide & provide & provide \\
\cellcolor{red!0.16}both & both & both \\
\cellcolor{red!0.13}analytical & analytical & analytical \\
\cellcolor{red!0.20}and & and & and \\
\cellcolor{red!0.07}technical & technical & technical \\
\cellcolor{red!0.20}support & support & support \\
\cellcolor{red!0.81}. & . & . \\
\bottomrule
\end{tabular}
\renewcommand{\arraystretch}{1.0} 
\end{table}

These results demonstrate that the token-level explanations produced by \contextmatch\ are both informative and, to a large extent, intuitively plausible to human experts.
Deeper analysis into potential improvements can be done in future work by extending the number of samples and annotators in this exercise.

\section*{Acknowledgment}

This project was funded by the Flemish Government, through Flanders Innovation \& Entrepreneurship (VLAIO, project HBC.2020.2893).
It also received funding from the Flemish government under the “Onderzoeksprogramma Artifici\"ele Intelligentie (AI) Vlaanderen” program.

\newpage

\bibliographystyle{unsrt}
\bibliography{refs}

\begin{thebibliography}{10}

\bibitem{dsjmaapplications}
Ibrahim Rahhal, Ismail Kassou, and Mounir Ghogho.
\newblock Data science for job market analysis: A survey on applications and techniques.
\newblock {\em Expert Systems with Applications}, 251:124101, 2024.

\bibitem{cunningham2016employer}
Wendy~V Cunningham and Paula Villase{\~n}or.
\newblock Employer voices, employer demands, and implications for public skills development policy connecting the labor and education sectors.
\newblock {\em The World Bank Research Observer}, 31(1):102--134, 2016.

\bibitem{9517309}
Imane Khaouja, Ismail Kassou, and Mounir Ghogho.
\newblock A survey on skill identification from online job ads.
\newblock {\em IEEE Access}, 9:118134--118153, 2021.

\bibitem{clavieskill}
Benjamin Clavi{\'{e}} and Guillaume Souli{\'{e}}.
\newblock Large language models as batteries-included zero-shot {ESCO} skills matchers.
\newblock In Mesut Kaya, Toine Bogers, David Graus, Chris Johnson, and Jens{-}Joris Decorte, editors, {\em Proceedings of the 3rd Workshop on Recommender Systems for Human Resources (RecSys in {HR} 2023) co-located with the 17th {ACM} Conference on Recommender Systems (RecSys 2023), Singapore, Singapore, 18th-22nd September 2023}, volume 3490 of {\em {CEUR} Workshop Proceedings}. CEUR-WS.org, 2023.

\bibitem{d2024context}
Karel D'Oosterlinck, Omar Khattab, Fran{\c{c}}ois Remy, Thomas Demeester, Chris Develder, and Christopher Potts.
\newblock In-context learning for extreme multi-label classification, 2024.

\bibitem{occcodingllm}
Parisa Safikhani, Hayastan Avetisyan, Dennis Föste-Eggers, and David Broneske.
\newblock Automated occupation coding with hierarchical features: a data-centric approach to classification with pre-trained language models.
\newblock {\em Discover Artificial Intelligence}, 3:6, 02 2023.

\bibitem{jobbert}
{Decorte, Jens-Joris and Van Hautte, Jeroen and Demeester, Thomas and Develder, Chris}.
\newblock {JobBERT : understanding job titles through skills}.
\newblock In {\em {FEAST, ECML-PKDD 2021 Workshop, Proceedings}}, page~{9}, {2021}.

\bibitem{Zhao_Javed_Jacob_McNair_2015}
Meng Zhao, Faizan Javed, Ferosh Jacob, and Matt McNair.
\newblock Skill: A system for skill identification and normalization.
\newblock {\em Proceedings of the AAAI Conference on Artificial Intelligence}, 29(2):4012--4017, Jan. 2015.

\bibitem{representationjobskill}
Shanshan Jia, Xiaoan Liu, Ping Zhao, Chang Liu, Lianying Sun, and Tao Peng.
\newblock Representation of job-skill in artificial intelligence with knowledge graph analysis.
\newblock In {\em 2018 IEEE Symposium on Product Compliance Engineering - Asia (ISPCE-CN)}, pages 1--6, 2018.

\bibitem{sayfullina}
Luiza Sayfullina, Eric Malmi, and Juho Kannala.
\newblock Learning representations for soft skill matching.
\newblock In Wil M.~P. van~der Aalst, Vladimir Batagelj, Goran Glava{\v{s}}, Dmitry~I. Ignatov, Michael Khachay, Sergei~O. Kuznetsov, Olessia Koltsova, Irina~A. Lomazova, Natalia Loukachevitch, Amedeo Napoli, Alexander Panchenko, Panos~M. Pardalos, Marcello Pelillo, and Andrey~V. Savchenko, editors, {\em Analysis of Images, Social Networks and Texts}, pages 141--152, Cham, 2018. Springer International Publishing.

\bibitem{zhang-etal-2022-skillspan}
Mike Zhang, Kristian Jensen, Sif Sonniks, and Barbara Plank.
\newblock {S}kill{S}pan: Hard and soft skill extraction from {E}nglish job postings.
\newblock In Marine Carpuat, Marie-Catherine de~Marneffe, and Ivan~Vladimir Meza~Ruiz, editors, {\em Proceedings of the 2022 Conference of the North American Chapter of the Association for Computational Linguistics: Human Language Technologies}, pages 4962--4984, Seattle, United States, July 2022. Association for Computational Linguistics.

\bibitem{zhang-etal-2024-nnose}
Mike Zhang, Rob van~der Goot, Min-Yen Kan, and Barbara Plank.
\newblock {NNOSE}: Nearest neighbor occupational skill extraction.
\newblock In Yvette Graham and Matthew Purver, editors, {\em Proceedings of the 18th Conference of the European Chapter of the Association for Computational Linguistics (Volume 1: Long Papers)}, pages 589--608, St. Julian{'}s, Malta, March 2024. Association for Computational Linguistics.

\bibitem{herandi2024skill}
Amirhossein Herandi, Yitao Li, Zhanlin Liu, Ximin Hu, and Xiao Cai.
\newblock Skill-llm: Repurposing general-purpose llms for skill extraction, 2024.

\bibitem{decorte2022design}
Jens-Joris Decorte, Jeroen Van~Hautte, Johannes Deleu, Chris Develder, and Thomas Demeester.
\newblock Design of negative sampling strategies for distantly supervised skill extraction, 2022.

\bibitem{gnehm-etal-2022-fine}
Ann-sophie Gnehm, Eva B{\"u}hlmann, Helen Buchs, and Simon Clematide.
\newblock Fine-grained extraction and classification of skill requirements in {G}erman-speaking job ads.
\newblock In David Bamman, Dirk Hovy, David Jurgens, Katherine Keith, Brendan O'Connor, and Svitlana Volkova, editors, {\em Proceedings of the Fifth Workshop on Natural Language Processing and Computational Social Science (NLP+CSS)}, pages 14--24, Abu Dhabi, UAE, November 2022. Association for Computational Linguistics.

\bibitem{decortellm}
{Decorte, Jens-Joris and Verlinden, Severine and Van Hautte, Jeroen and Deleu, Johannes and Develder, Chris and Demeester, Thomas}.
\newblock {Extreme multi-label skill extraction training using large language models}.
\newblock In {\em {AI4HR \& PES 2023 : International Workshop on AI for Human Resources and Public Employment Services, Proceedings}}, pages {1--10}, {2023}.

\bibitem{jiang2021lightxml}
Ting Jiang, Deqing Wang, Leilei Sun, Huayi Yang, Zhengyang Zhao, and Fuzhen Zhuang.
\newblock Lightxml: Transformer with dynamic negative sampling for high-performance extreme multi-label text classification.
\newblock In {\em Proceedings of the AAAI conference on artificial intelligence}, volume~35, pages 7987--7994, 2021.

\bibitem{dahiya2021deepxml}
Kunal Dahiya, Deepak Saini, Anshul Mittal, Ankush Shaw, Kushal Dave, Akshay Soni, Himanshu Jain, Sumeet Agarwal, and Manik Varma.
\newblock Deepxml: A deep extreme multi-label learning framework applied to short text documents.
\newblock In {\em Proceedings of the 14th ACM international conference on web search and data mining}, pages 31--39, 2021.

\bibitem{zhang2021fast}
Jiong Zhang, Wei-Cheng Chang, Hsiang-Fu Yu, and Inderjit Dhillon.
\newblock Fast multi-resolution transformer fine-tuning for extreme multi-label text classification.
\newblock {\em Advances in Neural Information Processing Systems}, 34:7267--7280, 2021.

\bibitem{hadsell2006dimensionality}
Raia Hadsell, Sumit Chopra, and Yann LeCun.
\newblock Dimensionality reduction by learning an invariant mapping.
\newblock In {\em 2006 IEEE computer society conference on computer vision and pattern recognition (CVPR'06)}, volume~2, pages 1735--1742. IEEE, 2006.

\bibitem{oord2018representation}
Aaron van~den Oord, Yazhe Li, and Oriol Vinyals.
\newblock Representation learning with contrastive predictive coding, 2018.

\bibitem{chen2020simple}
Ting Chen, Simon Kornblith, Mohammad Norouzi, and Geoffrey Hinton.
\newblock A simple framework for contrastive learning of visual representations.
\newblock In {\em International conference on machine learning}, pages 1597--1607. PmLR, 2020.

\bibitem{radford2021learning}
Alec Radford, Jong~Wook Kim, Chris Hallacy, Aditya Ramesh, Gabriel Goh, Sandhini Agarwal, Girish Sastry, Amanda Askell, Pamela Mishkin, Jack Clark, et~al.
\newblock Learning transferable visual models from natural language supervision.
\newblock In {\em International conference on machine learning}, pages 8748--8763. PmLR, 2021.

\bibitem{highprecphrase}
Ron Bekkerman and Matan Gavish.
\newblock High-precision phrase-based document classification on a modern scale.
\newblock In {\em Proceedings of the 17th ACM SIGKDD International Conference on Knowledge Discovery and Data Mining}, KDD '11, page 231–239, New York, NY, USA, 2011. Association for Computing Machinery.

\bibitem{carotene}
Faizan Javed, Qinlong Luo, Matt McNair, Ferosh Jacob, Meng Zhao, and Tae~Seung Kang.
\newblock Carotene: A job title classification system for the online recruitment domain.
\newblock In {\em 2015 IEEE First International Conference on Big Data Computing Service and Applications}, pages 286--293, 2015.

\bibitem{deepcarotene}
Jingya Wang, Kareem Abdelfatah, Mohammed Korayem, and Janani Balaji.
\newblock Deepcarotene -job title classification with multi-stream convolutional neural network.
\newblock In {\em 2019 IEEE International Conference on Big Data (Big Data)}, pages 1953--1961, 2019.

\bibitem{Zbib2022LearningJT}
Rabih Zbib, Lucas~Alvarez Lacasa, Federico Retyk, Rus Poves, Juan Aizpuru, Hermenegildo Fabregat, Vaidotas Šimkus, and Emilia García-Casademont.
\newblock {Learning Job Titles Similarity from Noisy Skill Labels}.
\newblock In {\em {FEAST, ECML-PKDD 2022 Workshop, Proceedings}}, {2022}.

\bibitem{VacancySBERT}
Maiia~Y. Bocharova, Eugene~V. Malakhov, and Vitaliy~I. Mezhuyev.
\newblock Vacancysbert: the approach for representation of titles and skillsfor semantic similarity search in the recruitment domain.
\newblock {\em Applied Aspects of Information Technology}, 6(1):52–59, April 2023.

\bibitem{laosaengpha-etal-2024-learning}
Napat Laosaengpha, Thanit Tativannarat, Chawan Piansaddhayanon, Attapol Rutherford, and Ekapol Chuangsuwanich.
\newblock Learning job title representation from job description aggregation network.
\newblock In Lun-Wei Ku, Andre Martins, and Vivek Srikumar, editors, {\em Findings of the Association for Computational Linguistics: ACL 2024}, pages 1319--1329, Bangkok, Thailand, August 2024. Association for Computational Linguistics.

\bibitem{gao-etal-2021-scaling}
Luyu Gao, Yunyi Zhang, Jiawei Han, and Jamie Callan.
\newblock Scaling deep contrastive learning batch size under memory limited setup.
\newblock In Anna Rogers, Iacer Calixto, Ivan Vuli{\'c}, Naomi Saphra, Nora Kassner, Oana-Maria Camburu, Trapit Bansal, and Vered Shwartz, editors, {\em Proceedings of the 6th Workshop on Representation Learning for NLP (RepL4NLP-2021)}, pages 316--321, Online, August 2021. Association for Computational Linguistics.

\bibitem{senger-etal-2024-deep}
Elena Senger, Mike Zhang, Rob van~der Goot, and Barbara Plank.
\newblock Deep learning-based computational job market analysis: A survey on skill extraction and classification from job postings.
\newblock In Estevam Hruschka, Thom Lake, Naoki Otani, and Tom Mitchell, editors, {\em Proceedings of the First Workshop on Natural Language Processing for Human Resources (NLP4HR 2024)}, pages 1--15, St. Julian{'}s, Malta, March 2024. Association for Computational Linguistics.

\bibitem{bhola-etal-2020-retrieving}
Akshay Bhola, Kishaloy Halder, Animesh Prasad, and Min-Yen Kan.
\newblock Retrieving skills from job descriptions: A language model based extreme multi-label classification framework.
\newblock In Donia Scott, Nuria Bel, and Chengqing Zong, editors, {\em Proceedings of the 28th International Conference on Computational Linguistics}, pages 5832--5842, Barcelona, Spain (Online), December 2020. International Committee on Computational Linguistics.

\bibitem{Shang2023}
Boyang Shang, Daniel~W. Apley, and Sanjay Mehrotra.
\newblock Diversity subsampling: Custom subsamples from large data sets.
\newblock {\em INFORMS Journal on Data Science}, 2(2):161–182, October 2023.

\bibitem{marchal-etal-2022-establishing}
Marian Marchal, Merel Scholman, Frances Yung, and Vera Demberg.
\newblock Establishing annotation quality in multi-label annotations.
\newblock In Nicoletta Calzolari, Chu-Ren Huang, Hansaem Kim, James Pustejovsky, Leo Wanner, Key-Sun Choi, Pum-Mo Ryu, Hsin-Hsi Chen, Lucia Donatelli, Heng Ji, Sadao Kurohashi, Patrizia Paggio, Nianwen Xue, Seokhwan Kim, Younggyun Hahm, Zhong He, Tony~Kyungil Lee, Enrico Santus, Francis Bond, and Seung-Hoon Na, editors, {\em Proceedings of the 29th International Conference on Computational Linguistics}, pages 3659--3668, Gyeongju, Republic of Korea, October 2022. International Committee on Computational Linguistics.

\bibitem{jensen-etal-2021-de}
Kristian~N{\o}rgaard Jensen, Mike Zhang, and Barbara Plank.
\newblock De-identification of privacy-related entities in job postings.
\newblock In Simon Dobnik and Lilja {\O}vrelid, editors, {\em Proceedings of the 23rd Nordic Conference on Computational Linguistics (NoDaLiDa)}, pages 210--221, Reykjavik, Iceland (Online), May 31--2 June 2021. Link{\"o}ping University Electronic Press, Sweden.

\bibitem{chalkidis-etal-2019-extreme}
Ilias Chalkidis, Emmanouil Fergadiotis, Prodromos Malakasiotis, Nikolaos Aletras, and Ion Androutsopoulos.
\newblock Extreme multi-label legal text classification: A case study in {EU} legislation.
\newblock In Nikolaos Aletras, Elliott Ash, Leslie Barrett, Daniel Chen, Adam Meyers, Daniel Preotiuc-Pietro, David Rosenberg, and Amanda Stent, editors, {\em Proceedings of the Natural Legal Language Processing Workshop 2019}, pages 78--87, Minneapolis, Minnesota, June 2019. Association for Computational Linguistics.

\bibitem{mpnet}
Kaitao Song, Xu~Tan, Tao Qin, Jianfeng Lu, and Tie-Yan Liu.
\newblock Mpnet: masked and permuted pre-training for language understanding.
\newblock In {\em Proceedings of the 34th International Conference on Neural Information Processing Systems}, NIPS '20, Red Hook, NY, USA, 2020. Curran Associates Inc.

\bibitem{SimCLR}
Ting Chen, Simon Kornblith, Mohammad Norouzi, and Geoffrey Hinton.
\newblock A simple framework for contrastive learning of visual representations.
\newblock In {\em Proceedings of the 37th International Conference on Machine Learning}, ICML'20. JMLR.org, 2020.

\end{thebibliography}

\begin{IEEEbiography}[{\includegraphics[width=1in,height=1.25in,clip,keepaspectratio]{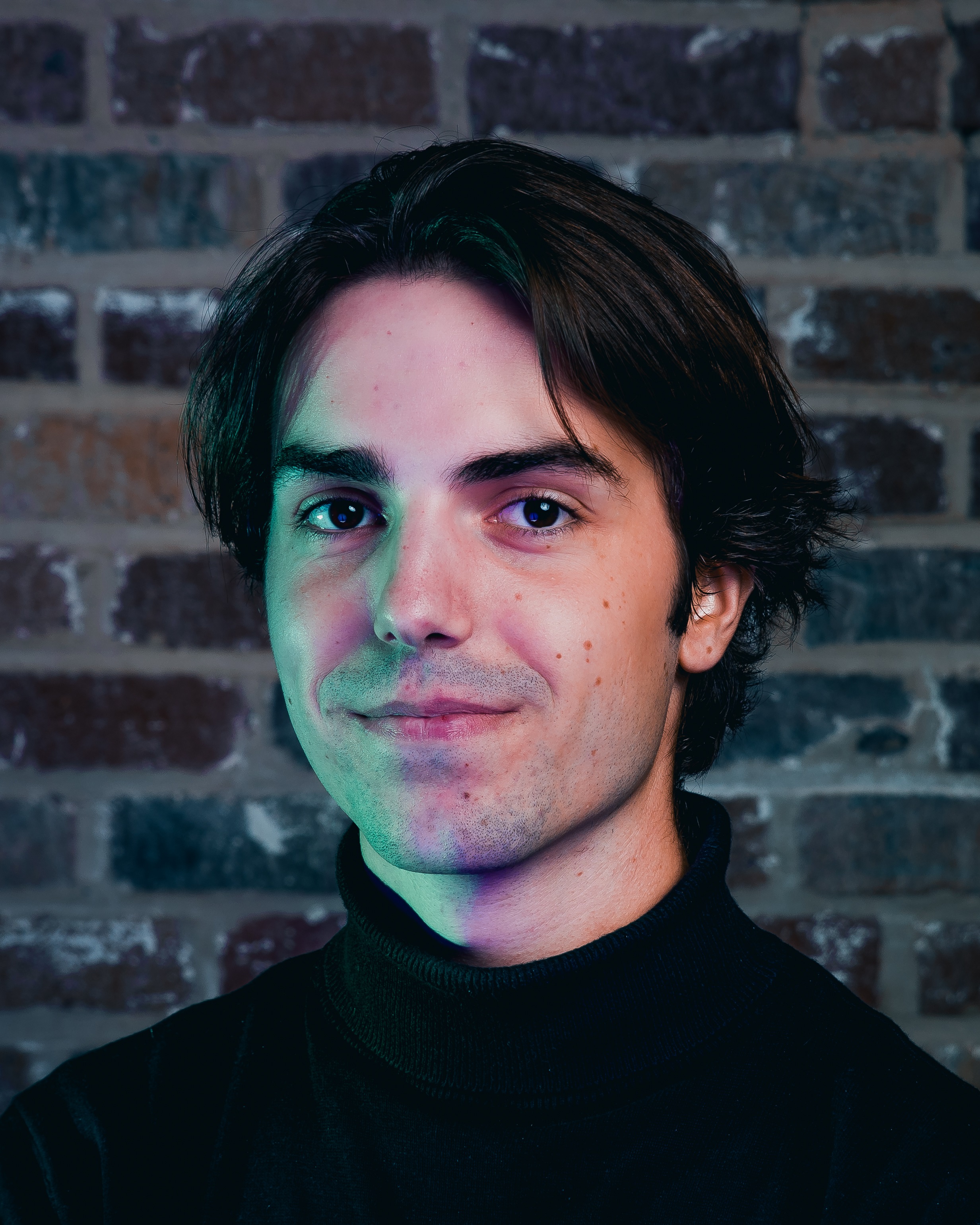}}]{Jens-Joris Decorte} received the M.S. degree in engineering from Ghent University, Belgium, in 2020. 
He is currently pursuing a Ph.D. degree with the Internet Technology and Data Science Laboratory (IDLab), Ghent University–imec, in collaboration with TechWolf. 
His supervisors are Prof. Chris Develder and Prof. Thomas Demeester. 
His research focuses on AI for HR applications, including skill extraction and skill-related tasks, using large-scale text encoding and contrastive learning methods.
He has been part of TechWolf since 2019, where he contributed to the development of the company’s AI systems and led the AI team. 
His research interests include natural language processing, skills intelligence, and machine learning.
In addition to his research and industry work, he is a lecturer in ICT evening education at Hogeschool Gent.
\end{IEEEbiography}

\begin{IEEEbiography}[{\includegraphics[width=1in,height=1.25in,clip,keepaspectratio]{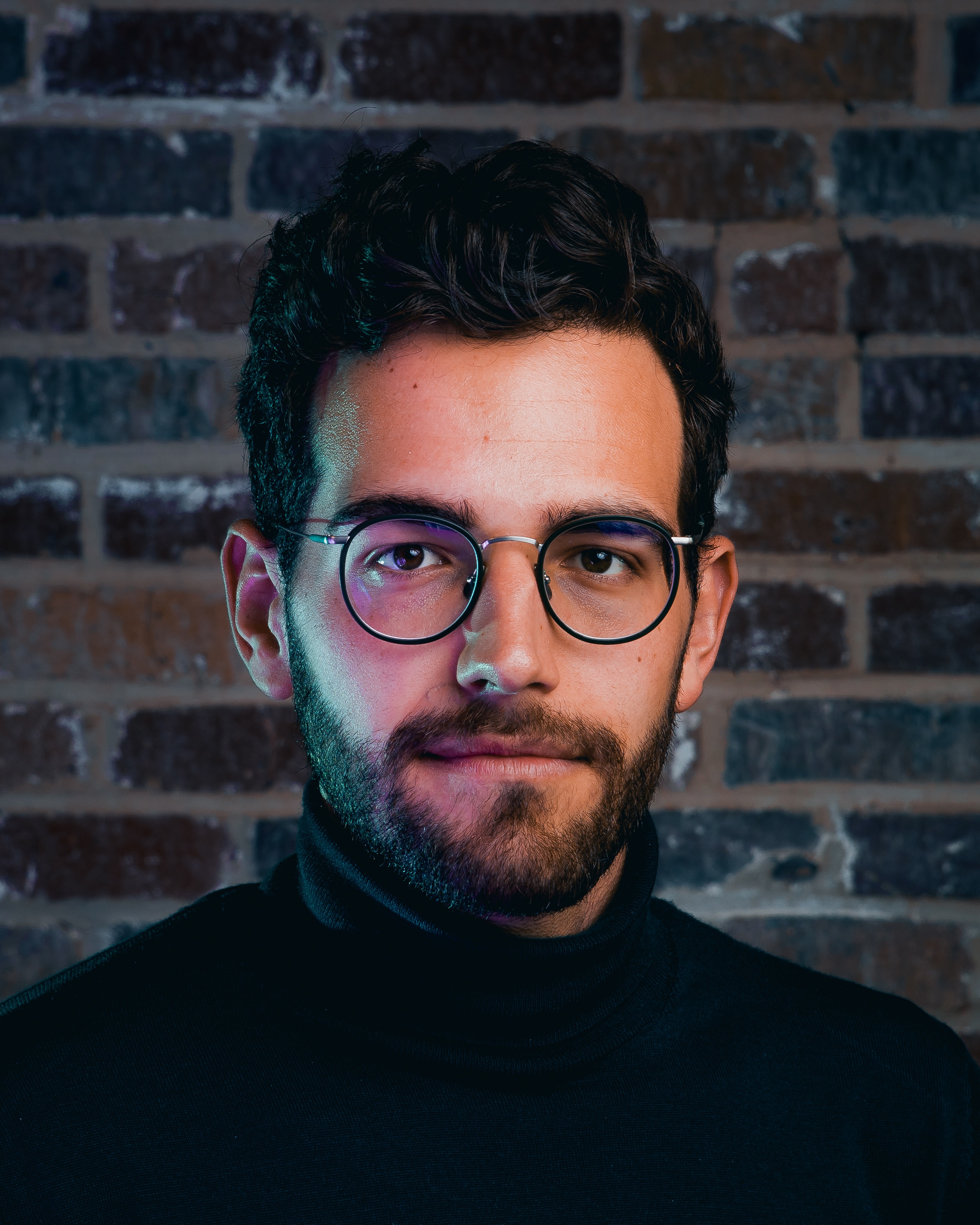}}]{Jeroen Van Hautte} 
is the co-founder and Chief Technology Officer of TechWolf, one of Europe's fastest-growing AI companies, specializing in skills intelligence. He received the M.Sc. degree in computer science engineering from Ghent University, Belgium, as well as the M.Sc. in advanced computer science from University of Cambridge. His AI research at the University of Cambridge led to a novel approach to capturing workforce skills.
His work in artificial intelligence and entrepreneurship has been widely recognized, including being named to the Forbes 30 Under 30 list and being named a Technology Pioneer by the World Economic Forum. At TechWolf, he oversees the product and engineering organization, as well as research and innovation.
\end{IEEEbiography}

\begin{IEEEbiography}[{\includegraphics[width=1in,height=1.25in,clip,keepaspectratio]{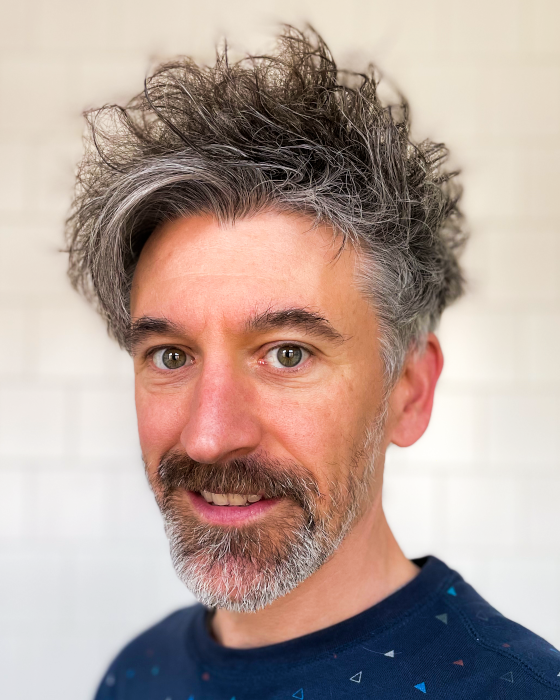}}]{Chris Develder} (Senior Member, IEEE) received the M.Sc. degree in computer science engineering and the Ph.D. degree in electrical engineering from Ghent University, Ghent, Belgium, in 1999 and 2003, respectively. 
He was a Research Visitor with the University of California at Davis, Davis, CA, USA, from July 2007 to October 2007, and Columbia University, New York, NY, USA, from January 2013 to June 2015. 
He is currently an Associate Professor with the Research Group IDLab, Department of Information Technology (INTEC), Ghent University–imec. 
He was/is involved in various national and European research projects, such as FP7 Increase, FP7 C-DAX, H2020 CPN, H2020 Bright, H2020 BIGG, H2020 RENergetic, and H2020 BD4NRG. 
He also (co-)leads two research teams within the IDLab, such as the UGent-T2K on converting text to knowledge, NLP, mostly information extraction using machine learning, and the UGent-AI4E on artificial intelligence for energy applications, smart grid. 
He has co-authored over 200 refereed publications in international conferences and journals. 
He is a fellow of the Research Foundation, FWO. 
He is a Senior Member of ACM and a member of ACL.
\end{IEEEbiography}

\begin{IEEEbiography}[{\includegraphics[width=1in,height=1.25in,clip,keepaspectratio]{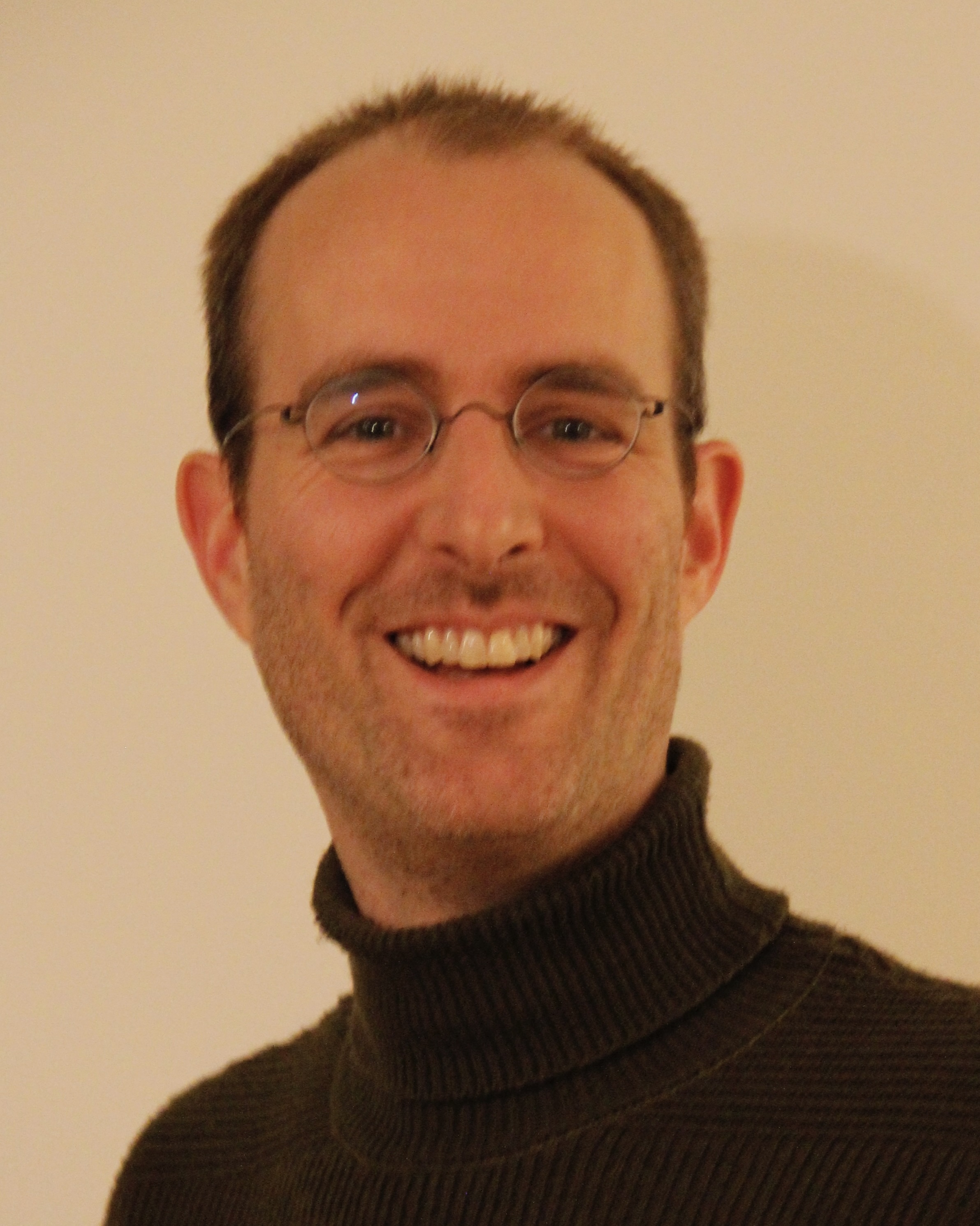}}]{Thomas Demeester} 
is an associate professor at IDLab, at the Department of Information Technology, Ghent University-imec in Belgium. After his master’s degree in electrical engineering (2005), he obtained his Ph.D. in computational electromagnetics, with a grant from the Research Foundation, Flanders (FWO-Vlaanderen) in 2009. His research interests then shifted to information retrieval (with a research stay at the University of Twente in The Netherlands, 2011),
natural language processing (NLP) and machine learning (with a stay at University College London in the UK, 2016), and
more recently to Neuro-Symbolic AI, with applications in AI for clinical and pre-clinical research. He has been involved in a series of national and international projects in the area of NLP, and co-authored around 150 peer-reviewed contributions in international journals and conferences. He is a member of AAAI and ELLIS.
\end{IEEEbiography}

\EOD

\end{document}